\documentclass[lettersize,journal]{IEEEtran}
\usepackage{amsmath,amsfonts}
\usepackage{algorithm}
\usepackage{algpseudocode}
\usepackage{array}
\usepackage[caption=false,font=normalsize,labelfont=sf,textfont=sf]{subfig}
\usepackage{textcomp}
\usepackage{stfloats}
\usepackage{url}
\usepackage{verbatim}
\usepackage{graphicx}
\usepackage{cite}
\usepackage{amsmath}
\usepackage{amssymb}
\usepackage{mathtools}
\usepackage{amsthm}

\usepackage{graphicx}
\usepackage{booktabs}
\usepackage{multirow}
\usepackage{orcidlink}

\usepackage[utf8]{inputenc} % allow utf-8 input
\usepackage[T1]{fontenc}    % use 8-bit T1 fonts
\usepackage{hyperref}       % hyperlinks
\usepackage{url}            % simple URL typesetting
\usepackage{booktabs}       % professional-quality tables
\usepackage{amsfonts}       % blackboard math symbols
\usepackage{nicefrac}       % compact symbols for 1/2, etc.
\usepackage{microtype}      % microtypography
\usepackage{xcolor}         % colors

\definecolor{red}{RGB}{0,0,0}

\begin{document} 

\title{View-Centric Multi-Object Tracking with Homographic Matching in Moving UAV}

\author{Deyi Ji$^{\orcidlink{0000-0001-7561-9789}}$, Lanyun Zhu$^{\orcidlink{0000-0001-7309-3330}}$, Siqi Gao, Qi Zhu$^{\orcidlink{0000-0002-1545-1854}}$, Yiru Zhao, Peng Xu,  Yue Ding$^{\orcidlink{0000-0002-2911-1244}}$, Hongtao Lu$^{\orcidlink{0000-0003-2300-3039}}$, \\ Jieping Ye$^{\orcidlink{0000-0001-8662-5818}}$,~\IEEEmembership{Fellow,~IEEE}, Feng Wu$^{\orcidlink{0000-0001-8451-0881}}$,~\IEEEmembership{Fellow,~IEEE}, and Feng Zhao$^{\orcidlink{0000-0001-6767-8105}}$,~\IEEEmembership{Member,~IEEE}
        % <-this % stops a space
\thanks{This work was supported by the Anhui Provincial Natural Science Foundation under Grant 2108085UD12, the Fundamental Research Funds for the Central Universities (project number YG2024ZD06), NSFC (No. 62176155),
Shanghai Municipal Science and Technology Major Project, China
(2021SHZDZX0102). (Corresponding author: Feng Zhao)}
\thanks{Deyi Ji, Qi Zhu, Feng Wu, and Feng Zhao are with MoE Key Laboratory of Brain-inspired Intelligent Perception and Cognition, University of Science and Technology of China.  (e-mail: jideyi@mail.ustc.edu.cn, zqcrafts@mail.ustc.edu.cn,  fengwu@ustc.edu.cn, fzhao956@ustc.edu.cn).}% <-this % stops a space
\thanks{Lanyun Zhu is with Singapore University of Technology and Design (e-mail: lanyun\_zhu@mymail.sutd.edu.sg).}
\thanks{Siqi Gao, Yiru Zhao, Peng Xu, and Jieping Ye are with Alibaba Group (e-mail: siqi.gsq@alibaba-inc.com, yiru.zyr@alibaba-inc.com, xupengcoding@gmail.com, jieping@gmail.com).}
\thanks{Yue Ding and Hongtao Lu are with Department of Computer Science and Engineering, Shanghai Jiao Tong University (e-mail: dingyue@sjtu.edu.cn, htlu@sjtu.edu.cn).}

}

\markboth{Journal of \LaTeX\ Class Files,~Vol.~14, No.~8, August~2021}%
{Shell \MakeLowercase{\textit{et al.}}: A Sample Article Using IEEEtran.cls for IEEE Journals}


\maketitle

\begin{abstract}
 In this paper, we address the challenge of Multi-Object Tracking (MOT) in moving Unmanned Aerial Vehicle (UAV) scenarios, where irregular flight trajectories, such as hovering, turning left/right, and moving up/down, lead to significantly greater complexity compared to fixed-camera MOT. Specifically, changes in the scene background not only render traditional frame-to-frame object IoU association methods ineffective but also introduce significant view shifts in the objects, which complicates tracking. To overcome these issues, we propose a novel HomView-MOT framework, which for the first time, harnesses the view homography inherent in changing scenes to solve MOT challenges in moving environments, incorporating homographic matching and view-centric concepts. We introduce a Fast Homography Estimation (FHE) algorithm for rapid computation of homography matrices between video frames, enabling object View-Centric ID Learning (VCIL) and leveraging multi-view homography to learn cross-view ID features. Concurrently, our Homographic Matching Filter (HMF) maps object bounding boxes from different frames onto a common view plane for a more realistic physical IoU association. Extensive experiments have proven that these innovations allow HomView-MOT to achieve state-of-the-art performance on prominent UAV MOT datasets VisDrone and UAVDT.
\end{abstract}

\begin{IEEEkeywords}
Multi-Object Tracking, UAV, Homographic Matching.
\end{IEEEkeywords}

\section{Introduction}
\label{sec:intro}

\IEEEPARstart{M}{ulti}-Object Tracking (MOT) has been a classical and longstanding task in various wide applications \cite{app1,app2}.
It is aimed to address the cross-frame trajectory of each object in video frames. Various MOT paradigms have been proposed for conventional static scene captured by fixed cameras \cite{sort, u2mot, motr}. Recently, benefited from the advancement of Unmanned Aerial Vehicle (UAV) technologies \cite{dlpl,uavmot,pptformer}, the accessibility and analysis of scenes under moving UAV views is opening new horizons for the community \cite{uavmot,visdrone}.

\begin{figure}
    \centering
    \includegraphics[width=\linewidth]{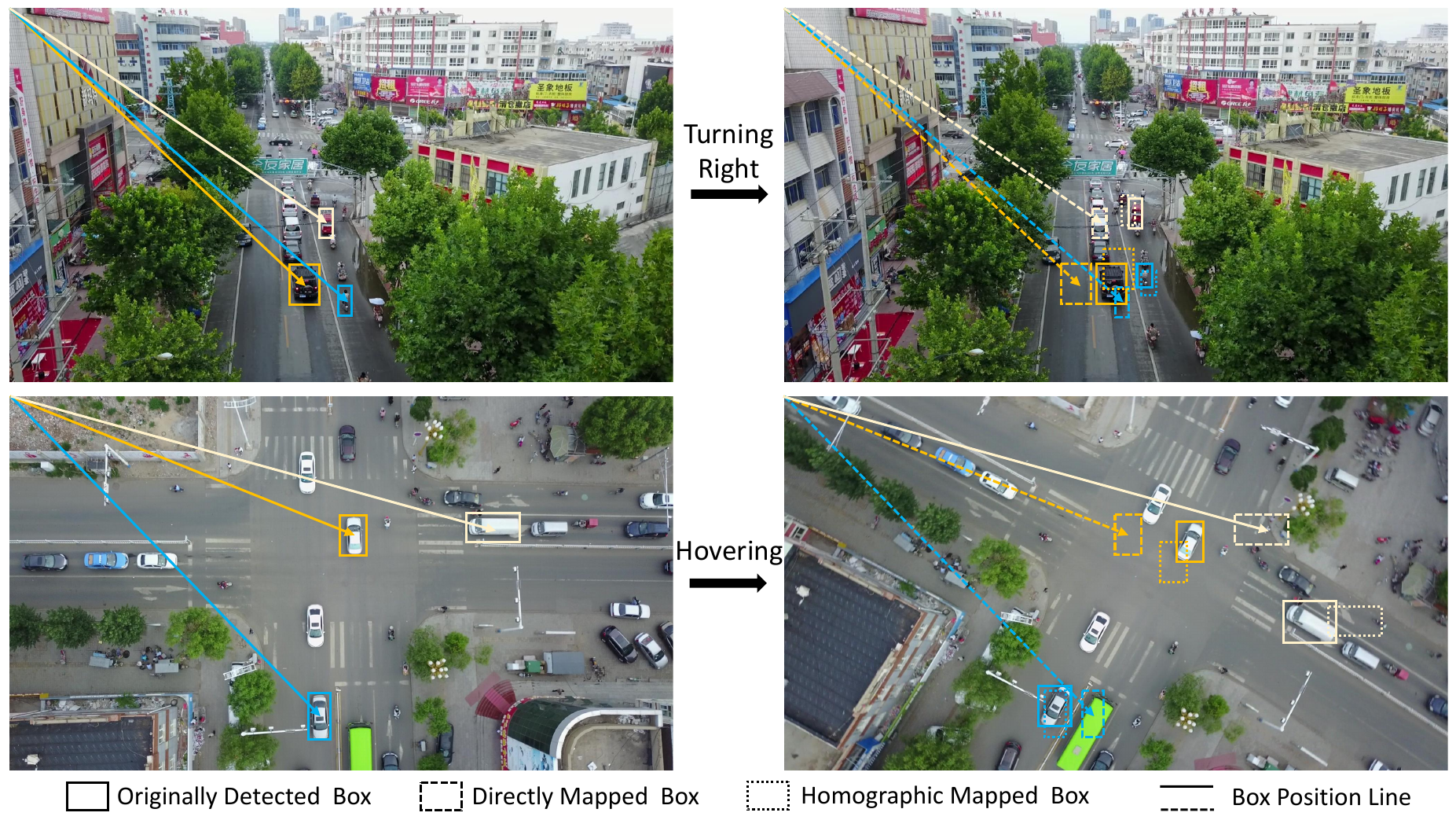}
    \caption{Tracking failures caused by irregular UAV flight states (turning right, hovering) leading to the ineffectiveness of ordinary IoU association. For convenience, we directly map the object box coordinates from the previous frame to the next frame (Box Position Line for clearly illustrating this process). As seen the IoU between the box of the same object across two frames (Directly Mapped Box of the previous frame and Originally Detected Box in the next frame) is very small (may be zero), causing tracker to easily lose tracking. The proposed Homographic Matching Filter utilizes homography spatial relationship between two frames to map the previous frame box homographically onto the next frame before performing IoU association (Homographic Mapped Box of the previous frame, and Originally Detected Box in the next frame), clearly showing a substantial increase and a more reasonable IoU (physical IoU in the real world).}
    \label{fig:intro_box}
\end{figure}

\begin{figure}
    \centering
    \includegraphics[width=1\linewidth]{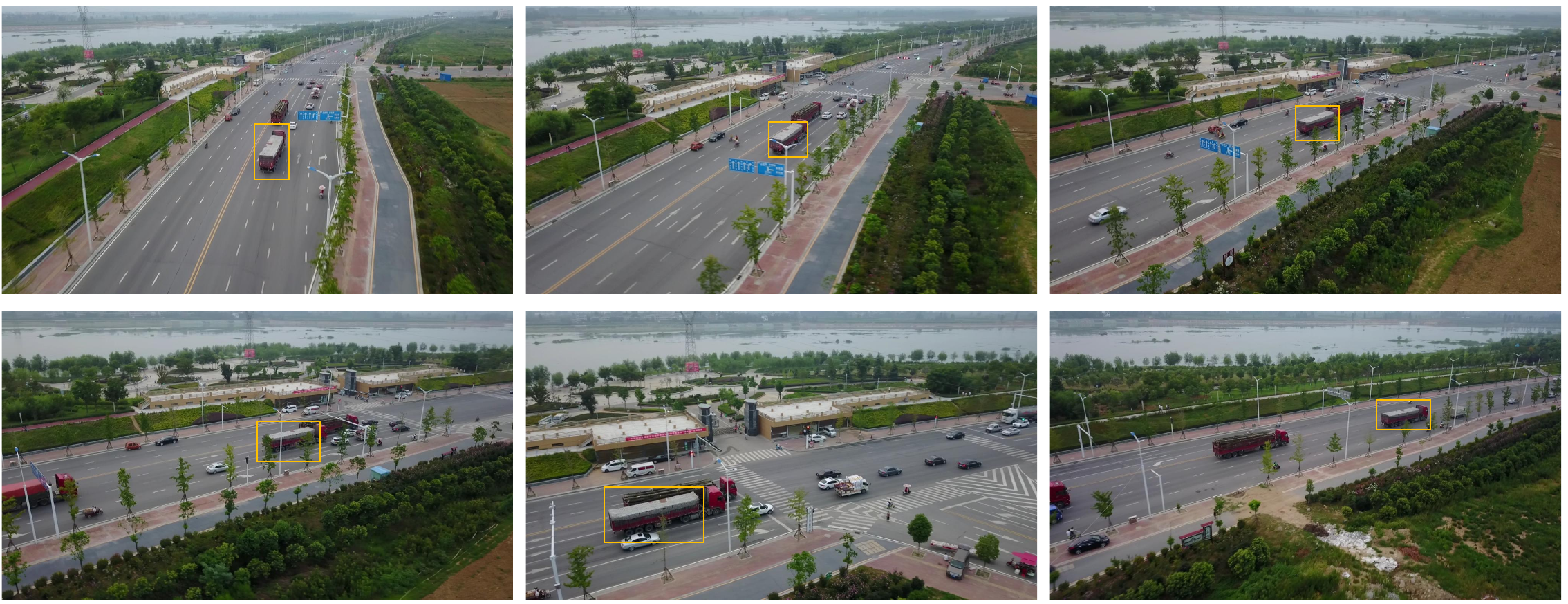}
    \caption{The irregular flight trajectory of the UAV causes the captured objects to exhibit a richer variety of view changes relative to a fixed shooting scene, posing greater challenges for MOT algorithms.}
    \label{fig:intro_view}
\end{figure}

Compared to the fixed scenes captured by conventional stationary cameras, due to the movement of the UAV itself in a moving UAV scenario, the captured scene is also in motion, making it more difficult to track multiple objects with different motion states within the scene.
Specifically, this manifests in two main ways: (1) During the irregular trajectories in the UAV flight, the variable shooting angles of the UAV lead to a rich diversity of view changes in the captured objects, posing a significant challenge for object ID discrimination learning during the tracking process, as shown in Figure \ref{fig:intro_view}.
(2) As the overall scene changes, it becomes challenging to capture the actual motion states of the objects, leading to numerous tracking losses. Specifically, as the UAV moves, the relative coordinates of the objects against the background become highly variable, introducing significant errors and uncertainties in the frame-to-frame ID matching process during tracking.  As shown in the examples in Figure \ref{fig:intro_box}, we can see that the vehicles framed in two consecutive frames have only moved slightly relative to the ground in the real world. If captured by a fixed camera with an unchanging scene, these objects could be easily matched through IoU association, which is widely used in classical MOT algorithms. However, due to the UAV's potential for various state changes such as hovering (row 2 in Figure \ref{fig:intro_box}),  turning left/right (row 1 in Figure \ref{fig:intro_box}), and moving up/down, the IoU of the absolute coordinates between the two frames (calculated between the directly mapped box of previous frame and the originally detected box in next frame) can be very small (even zero), resulting in match losses.

In this paper, we envision utilizing homography property of spatial scenes to address the aforementioned issues and propose a novel \textbf{View}-Centric Multi-Object Tracking with \textbf{Homographic} Matching (HomView-MOT) in Moving UAV scene. Firstly, for each UAV-captured video sequence, we regard it as multi-view shooting of one or more scenes. Therefore, we initially compute the homography transformation matrix between adjacent frames. Notably, to enhance tracking efficiency, we avoid calculating between every two adjacent frames and instead perform skip-frame computation, only calculating the homography matrix for every $h$ frames apart. We design an efficient Fast homography Estimation (FHE) algorithm to reasonably estimate the homography matrix for every two key frames within this interval. Secondly, based on the inter-frame scene homography matrix, we present the Homographic Matching Filter (HMF), which can map detected objects within them to the same viewpoint space to calculate the "physical IoU in the real world" of objects between two frames (as shown in Figure \ref{fig:intro_box}), thereby improving the precision of track matching. Finally, to address the issue of object view changes during UAV flight, we propose View-Centric ID Learning (VCIL). 
By combining the inter-frame scene homography matrix, we design Homographic Slot Attention (HSA) to explicitly extract view-centric slots for each frame and correlate views between adjacent frames, thus updating object ID features to cope with view changes during UAV flight. By integrating these designs, we make HomView-MOT an efficient and effective multi-object tracker.

Essentially, HomView-MOT is a specialized framework designed for "multi-object tracking in moving scenarios". In this paper, we validate our approach on two typical UAV MOT datasets, including VisDrone 2019 \cite{visdrone} and UAVDT \cite{uavdt}. 
The main contributions of this paper are as follows:

\begin{itemize}

\item For the first time, we address the problem of Multi-Object Tracking (MOT) in moving UAV scenarios by integrating the concept of homography from different views, introducing Homographic Matching and View-Centric concepts into the field and proposing a universal HomView-MOT framework.

\item Specifically, we begin by employing a Fast Homography Estimation (FHE) algorithm to rapidly estimate the homography matrix between any two key frames. Based on this, we perform object View-Centric ID Learning (VCIL), utilizing the homography of multiple perspectives to learn cross-view ID features. Furthermore, leveraging the homography matrix, we design a Homographic Matching Filter (HMF) to map the object bounding boxes from different frames onto the same view plane, allowing for more accurate physical IoU associations that closely mirror the real world.

\item Through extensive experiments, we validate that the above designs enable HomView-MOT to achieve state-of-the-art performances on two typical UAV MOT datasets, VisDrone and UAVDT.

\end{itemize}

\begin{figure*}
    \centering
    \includegraphics[width=0.8\linewidth]{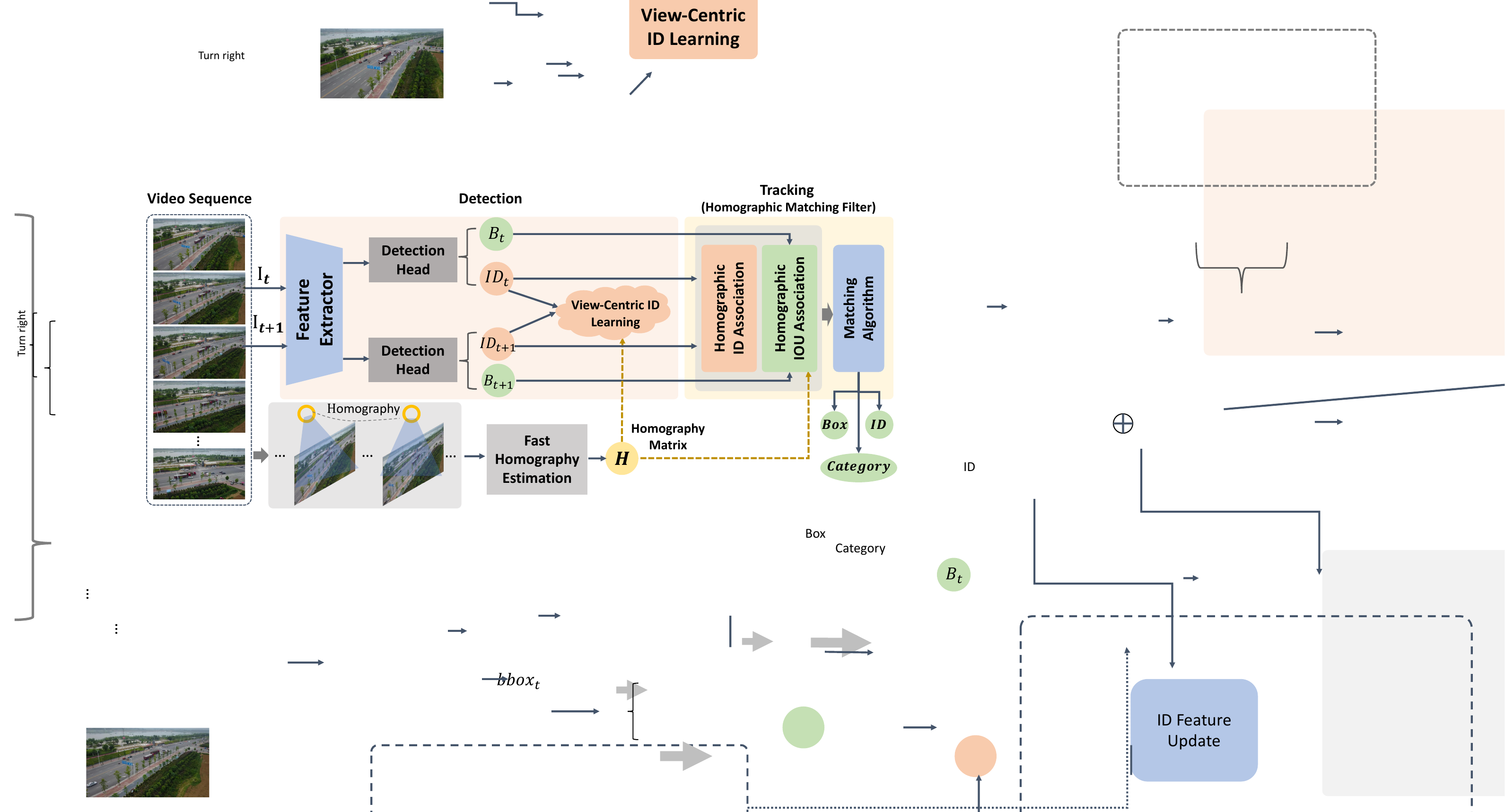}
    \caption{The overview of our proposed HomView-MOT.
    It begins with a Fast Homography Estimation (FHE) algorithm that rapidly calculates the homography transformation matrix between any two adjacent frames for view scene projection.
    Subsequently, the View-Centric ID Learning (VCIL) is utilized to explicitly assimilate the multi-view information and update the ID features. For tracking, HMF is employed to project the object boxes from adjacent frames onto a common view planar space for IoU association. This is coupled with ID association and processed through  Matching Algorithm to produce final tracking results.}
    \label{fig:overview}
\end{figure*}

\section{Related Work}

\subsection{Multi-Object Tracking}

Classical MOT algorithms generally adopt a tracking by detection framework \cite{sort,dsort,ocsort,han2023mmptrack,urur,yang2023integrating}, where the first step is to detect all object boxes in each frame in videos\cite{zhu2025skysense,tgrs11,raven}, and the second step involves object ID matching between adjacent frames through data association\cite{bytetrack,dsort}. In the first detection step, related algorithms such as SORT\cite{sort}, Deep SORT\cite{dsort}.
Recent research has also introduced vision transformer detectors into MOT to achieve higher performances, as seen in TransTrack\cite{transtrack} and TrackFormer\cite{trackformer}.
In the second step of data association, traditional MOT algorithms usually rely on Matching Algorithms based on IoU association between target boxes in adjacent frames. This method is quite effective in static scenes captured by fixed cameras. However, in UAV footage and the like, the moving scene background, causing different frames to be on different viewing planes \cite{dlpl,tgrs8,pptformer}, can render the conventional IoU association method inaccurate. To address this, UAVMOT\cite{uavmot} proposes an Adaptive Motion Filter to judge the different motion states of the UAV and process them differently. 
{\color{red}Moreover, ASTCA \cite{yuan2022learning} incorporates spatial-temporal context to mitigate background interference and the boundary effect in UAV tracking, which is a challenge also pertinent to MOT in complex aerial views. UDALT \cite{liang2025uncertainty} offers a fascinating perspective on reducing annotation costs by selecting the most informative samples. Self-SDCT \cite{yuan2020self} utilize multi-cycle consistency loss as self-supervised signal to learn feature representations from adjacent frames without manual annotations is highly relevant.}

\subsection{ID Embedding Assistance and View-Oriented Learning}

To enhance robustness, many Multi-Object Tracking (MOT) frameworks propose adding an embedding based on ReID (re-identification) learning \cite{tgrs1,sstkd_pami,tgrs5,tgrs6,cagcn,zhu2025not,zhuretrv} on top of detection, which is used to improve multi-object data association \cite{fairmot,sstkd,uavmot} in videos \cite{tgrs7,raven++,tgrs9}. For instance, FairMOT \cite{fairmot} integrates joint detection and ReID learning into a unified framework, achieving state-of-the-art performance in pedestrian tracking tasks. Similarly, UAVMOT \cite{uavmot} focuses on addressing challenges specific to aerial views, such as scale variations and occlusions, by leveraging robust ReID embeddings.
In scenes captured by stationary cameras, targets generally present a very stable viewpoint, making ID learning comparatively simple and straightforward. This stability allows traditional methods like DeepSORT \cite{deepsort} to rely heavily on appearance features for accurate data association. However, in the case of UAVs (Unmanned Aerial Vehicles), due to the variability of shooting angles, an object may exhibit a rich array of viewpoints. This complexity introduces significant challenges for conventional ReID approaches, as highlighted in studies like \cite{zheng2019joint,chen2019abd,gpwformer}, which emphasize the need for viewpoint-invariant representations in aerial scenarios.

{\color{red}{\subsection{Camera Motion Compensation}

The challenge of camera motion in Multi-Object Tracking (MOT) has been actively addressed by Camera Motion Compensation (CMC) techniques. For instance, UCMCTrack \cite{ucmctrack} introduces a uniform compensation model and a ground plane-based Mapped Mahalanobis Distance, demonstrating robustness in various scenarios but potentially struggling with the non-uniform, irregular motion patterns characteristic of UAV flights. Similarly, another line of research focuses on Global Motion Compensation (GMC) for UAVs \cite{zhang2024reliable,mojidra2023vision}, often employing frame-by-frame feature matching to align consecutive images and mitigate the impact of camera movement. While these methods effectively stabilize the scene, their computational cost can be high, and they primarily address the background motion at the association level. In contrast, our HomView-MOT framework advances the CMC paradigm by introducing a highly efficient Fast Homography Estimation (FHE) algorithm that uses keyframes and interpolation to reduce computational burden significantly. }}

\section{Method}

\subsection{Overview}

Our proposed HomView-MOT, as shown in Figure \ref{fig:overview}, follows the classic tracking-by-detection paradigm. Given a video sequence acquired by a moving UAV, the goal of MOT is to obtain the categories, bounding boxes, and tracking IDs of objects in the video. First of all, we present a Fast Homography Estimation (FHE) algorithm to calculate the homography transformation matrix for any adjacent frames, which is used for scene view projection. Taking the adjacent $t$-th and $t+1$-th frames as an example, these two frames complete the detection of objects through the detection head of a shared feature extraction network and output bounding boxes (specifically, obtained by the heatmap and the width and height of the box), as well as objects ID embedding features. Next, we propose a View-Centric ID Learning (VCIL) module to explicitly learn the multi-view features of each object and update the object ID features. Then, in the tracking phase, we propose the homography Matching Filter (HMF), which maps the object boxes from adjacent frames onto the same view planar space for IoU association, combined with ID feature similarity, and finally sends them to the Matching Algorithm to output the final tracking results.

\subsection{Fast Homography Estimation for View Projection}

\begin{figure}
    \centering
    \includegraphics[width=1\linewidth]{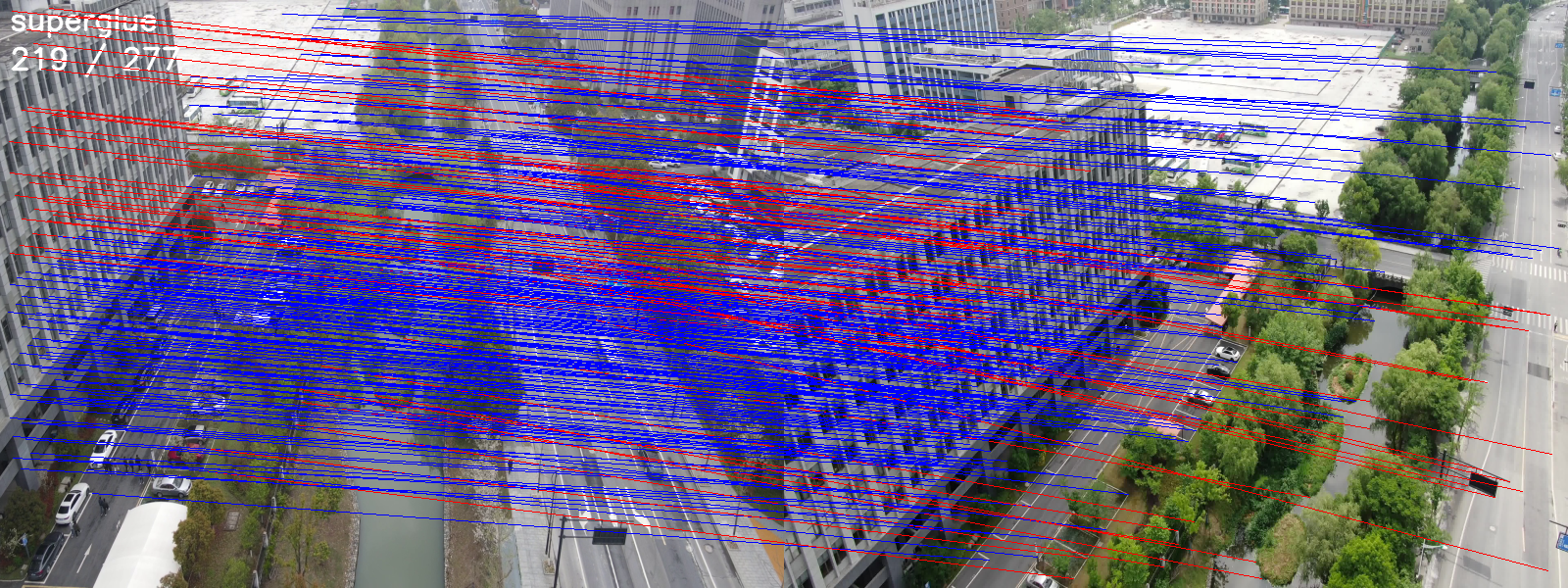}
    \caption{One example of the calculation of homography matrix. The lines indicate the matching of keypoints between the two frames.}
    \label{fig:view_projection}
\end{figure}

As described in Sec. \ref{sec:intro}, when the UAV flight state changes (e.g. hovering, moving up/down, turning left/right) occur, the captured scene background will undergo significant changes, making adjacent frames appear as different view planes over the same scene. To facilitate cross-view learning and IoU association for object IDs, we propose to perform view projection for these two planes, in the form of homography matrix which is extensively applied to various 3D applications \cite{homography_3d}. Generally, the homography matrix for every two adjacent frames can be calculated with the Homography Estimation (HE) algorithm. For instance, as shown in Figure \ref{fig:view_projection}, we use a shared SuperPoint \cite{superpoint} network to extract keypoints from both frames separately. The distribution of these keypoints is able to represent the view information of the scene \cite{zeng2018rethinking,liu2023geometrized,zhang19963d,li2020srhen}. Then, we can estimate the homography matrix based on the matching of keypoints between the two frames by the SuperGlue algorithm \cite{superglue}.

However, directly performing homography estimation calculation for every two frames is very time-consuming and would greatly reduce tracking efficiency. Therefore, we propose a Fast Homography Estimation (FHE) method, where we only need to directly calculate the homography matrices for a limited number of key frames, while the homography matrices for other adjacent frames can be fast estimated using the ones of the key frames. Specifically, we utilize a uniform sampling process to select the key frames from the input video sequence $I_t (t \in [1, T])$, the sampling interval is set to $h$. We only perform a direct calculation of the homography matrices $\{H_{t,t+h}, H_{t+h, t+2h}, ...\}$ between the sampled frames, where $H_{t, t+h}$ indicates the homography matrix from the $t+h$-th to $t$-th frame. For any two frames, based on the following Preliminary of Homography Estimation, we prove that their homography matrix can be reasonably estimated through the derivation of homography matrix of their adjacent key frames.

\subsection{Derivation of Homography Estimation in Adjacent Frames}

When the sampling interval $h$ is sufficiently small, the UAV can be approximated as moving in a nearly linear fashion within this interval. Taking the sampling frames $k_1, k_2 (k_2=k_1+h)$ and any non-sampling frame $t (k_1<t<k_2)$ as an example, for any two points $(x_{k_1}, y_{k_1})$ and $(x_{t}, y_{t})$ on the $t$ plane and $k_1$ plane respectively, it is approximately true that 

\begin{equation}
\begin{aligned}
    x_t = \alpha_1 x_{k_2}, ~~ y_t = \alpha_2 y_{k_2},    
\end{aligned}
\label{eq1}
\end{equation}
where $\alpha_1, \alpha_2$ are scaling factors based on the proportion of the overall displacement of key points between the two frames to the overall displacement of key points between the adjacent key frames $k_1, k_2$. 
Formally, let $M_{t, k_1}=\{(p_{t}^1, p_{k_1}^1), (p_{t}^2, p_{k_1}^2),..., (p_{t}^z, p_{k_1}^z), ..., (p_{t}^{Z_{t,k_1}}, p_{k_1}^{Z_{t,k_1}})\}$ denote the matched keypoints between $t$ plane and $k_1$ plane, obtained by the SuperPoint + SuperGlue algorithm in Section 3.2 in the main paper. $(p_{t}^{z}, p_{k_1}^{z})$ indicate the $z$-th matched keypoints in $t$ plane and $k_1$ plane respectively. $Z_{t,k_1}$ indicates the maximum number of matched point pairs. $p^z_{k_1}=(x^z_{k_1}, y^z_{k_1})$, where $x^z_{k_1}, y^z_{k_1}$ indicate the $x$ and $y$ coordinate of $p^z_{k_1}$ respectively.

Then, $\alpha_1, \alpha_2$ are scaling factors based on the proportion of the overall displacement of key points between $t$ and $k_1$ plane to the overall displacement of key points between the adjacent key frames $k_1, k_2$. Formally, 

\begin{equation}
\begin{aligned}
\alpha_1 &= \frac{\frac{1}{Z_{t,k_1}}\sum_{z=1}^{Z_{t,k_1}} D(x_t^z, x_{k_1}^z)}{\frac{1}{Z_{k_2,k_1}}\sum_{z=1}^{Z_{k_2,k_1}} D(x_{k_2}^z, x_{k_1}^z)} , \\
\alpha_2 &= \frac{\frac{1}{Z_{t,k_1}}\sum_{z=1}^{Z_{t,k_1}} D(y_t^z, y_{k_1}^z)}{\frac{1}{Z_{k_2,k_1}}\sum_{z=1}^{Z_{k_2,k_1}} D(y_{k_2}^z, y_{k_1}^z)} ,
\end{aligned}
\label{eq11}
\end{equation}
\noindent where $D(\cdot)$ indicates the L2 distance.

Finally, $\alpha_{t, k_1}$ can be calculated by Eq. \ref{eq8} and Eq. \ref{eq11}. According to the homography property \cite{homography}, the $t$ plane and $k_1$ plane represent different views of the same scene, so we have
\begin{equation}
\begin{aligned}
x_t=H_{t,k_1}x_{k_1}, ~~ y_t=H_{t,k_1}y_{k_1}.
\end{aligned}
\label{eq2}
\end{equation}

\begin{figure}
\centering
  % \begin{minipage}[c]{0.5\textwidth}
    \includegraphics[width=0.35\textwidth]{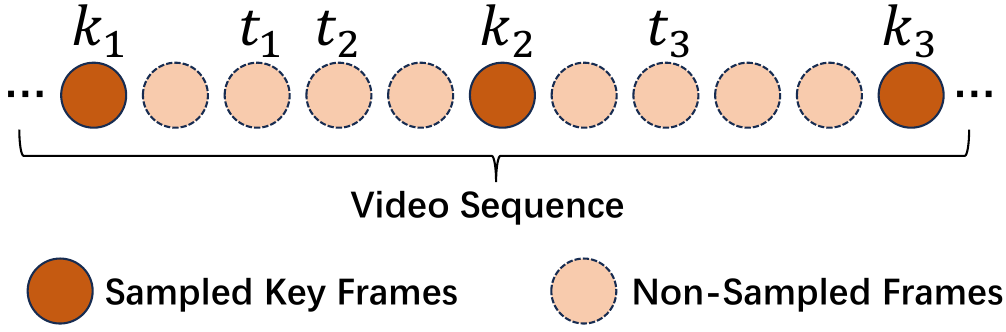}
  % \end{minipage}\hfill
  % \begin{minipage}[c]{0.45\textwidth}
    \caption{
       Illustration of the sampled key frames and non-sampled frames in the video sequence, where $h=5$, for example.
    } \label{fig:fhe}
  % \end{minipage}
\end{figure}

From the derivation of homography estimation \cite{homography, homography_3d} algorithm, we know that
\begin{equation}
\begin{aligned}
    A_{t,k_1}H_{t, k_1}=0,
\end{aligned}
\label{eq3}
\end{equation}

\noindent where

\begin{equation}
    \resizebox{0.9\hsize}{!}{$A_{t,k_1} = \left[
                                \begin{aligned}
                                 & ~ -x_{k_1} ~ -y_{k_1} ~~ -1 ~~~~ 0 ~~~~~~~~ 0  ~~~~~~~~ 0 ~~~ x_tx_{k_1} ~~ x_ty_{k_1} ~~ x_t~ \\
                                 & ~~~~ 0 ~~~~~~~ 0 ~~~~~~~~ 0 ~~ -x_{k_1} ~~ -y_{k_1} ~~ -1 ~~ y_tx_{k_1} ~~ y_ty_{k_1} ~~~ y_t~
                                \end{aligned}
                                \right].$}
\label{eq4}
\end{equation}

\noindent Similarly, we have

\begin{equation}
\begin{aligned}
A_{k_2,k_1}H_{k_2, k_1}=0,
\end{aligned}
\label{eq5}
\end{equation}

\noindent where 

\begin{equation}
% \begin{aligned}
\resizebox{0.9\hsize}{!}{$A_{k_2,k_1} = \left[
                                \begin{aligned}
                                 & ~ -x_{k_1} ~ -y_{k_1} ~~ -1 ~~~~ 0 ~~~~~~~~ 0  ~~~~~~~~ 0 ~~~ x_{k_2}x_{k_1} ~~ x_{k_2}y_{k_1} ~~ x_{k_2}~ \\
                                 & ~~~~ 0 ~~~~~~~ 0 ~~~~~~~~ 0 ~~ -x_{k_1} ~~ -y_{k_1} ~~ -1 ~~ y_{k_2}x_{k_1} ~~ y_{k_2}y_{k_1} ~~~ y_{k_2}~
                                \end{aligned}
                                \right]. $}
% \end{aligned}
\label{eq6}
\end{equation}

\noindent Through the Eq. \ref{eq4}, Eq. \ref{eq6} and Eq. \ref{eq1}, we can derive that

\begin{equation}
\begin{aligned}
A_{t,k_1}=\alpha_{t,k_1}A_{k_2,k_1},
\end{aligned}
\label{eq7}
\end{equation}

\noindent where 

\begin{equation}
\begin{aligned}
\alpha_{t,k_1} = \left[
                                \begin{aligned}
                                 & ~ 1 ~~ 1 ~~ 1 ~~ 1 ~~ 1  ~~ 1 ~~ \alpha_1 ~~ \alpha_1 ~~ \alpha_1~ \\
                                 & ~ 1 ~~ 1 ~~ 1 ~~ 1 ~~ 1 ~~ 1 ~~ \alpha_2 ~~ \alpha_2 ~~ \alpha_2 ~
                                \end{aligned}
                                \right].
\end{aligned}
\label{eq8}
\end{equation}

\noindent Next, through Eq. \ref{eq7}, Eq. \ref{eq3} and Eq. \ref{eq5}, we can get,

\begin{equation}
\begin{aligned}
A_{t,k_1}H_{t,k_1}=0=\alpha_{t,k_1}A_{k_2,k_1}H_{k_2,k_1},
\end{aligned}
\label{eq9}
\end{equation}
\noindent finally, we can derive that,

\begin{equation}
\begin{aligned}
H_{t,k_1}=\alpha_{t,k_1}H_{k_2,k_1}.
\end{aligned}
\label{eq10}
\end{equation}

% \noindent Proof completed.

\begin{figure}
\centering
% \begin{minipage}[c]{0.7\textwidth}
\includegraphics[width=1\linewidth]{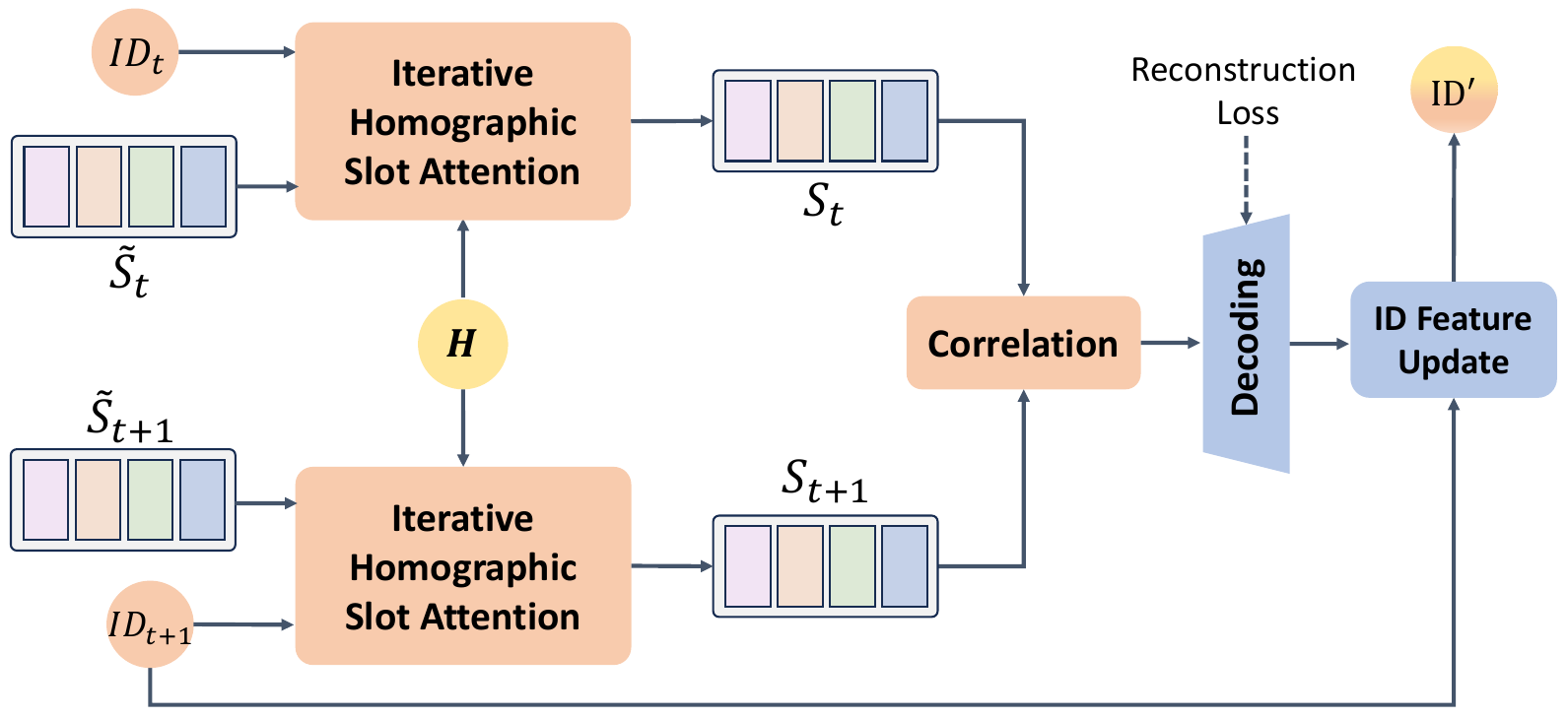}
% \end{minipage}\hfill
% \begin{minipage}[c]{0.28\textwidth}
\caption{
   The illustration of View-Centric ID Learning. The $\widetilde{S_t}$ and $\widetilde{S}_{t+1}$ indicate the randomly initialized view-centric slots.
} 
\label{fig:vcil}
% \end{minipage}
\end{figure}

\subsection{Efficient Video-Level Homography Transformation}

Specifically, as shown in Figure \ref{fig:fhe}, We select three highly representative frames $I_{t_1}, I_{t_2}, I_{t_3}$ to illustrate the computation process of the homography matrix between any two frames in the input video, where $k_1 < t_1 < t_2 < k_2 < t_3 < k_3$, and $k_1, k_2, k_3$ are the sampled frames. The core calculations are as follows:
\begin{enumerate}
 \item  For the same frame, that is, for any given frame $t$, according to the property of homography, we have $H_{t,t}=E$, where $E$ is the identity matrix.
 \item  Between any two sampled key frames, we compute the homography matrix $H_{k_2, k_1}, H_{k_3, k_1}, H_{k_3, k_2}$ using the naive homography estimation algorithm.
 \item  For any two frames $t_1, t_2$ within the same sampling interval:
 \begin{itemize}
   \item From a non-sampled frame to the previous sampled frame: $H_{t_1, k_1}=\alpha_{t_1, k_1}H_{k_2,k_1}$ and $H_{t_2, k_1}=\alpha_{t_2, k_1}H_{k_2,k_1}$, where $\alpha$ is scaling factors based on the proportion of the overall displacement of key points between the two frames to the overall displacement of key points between the adjacent key frames $k_1, k_2$. More details refer to the appendix.
   \item From a sampled frame to a non-sampled frame: $H_{k_2, t_1}=H_{k_2,k_1}H^{-1}_{t_1,k_1}$ and $H_{k_2, t_2}=H_{k_2,k_1}H^{-1}_{t_2,k_1}$.
   \item From a non-sampled frame to another non-sampled frame: $H_{t_2,t_1}=H_{t_2,k_1}H_{t_1,k_1}^{-1}$.
 \end{itemize}
 \item  For any two frames $t_1, t_3$ in different sampling intervals:
 \begin{itemize}
   \item From a non-sampled frame to any frame in the previous sampling interval: $H_{t_2,t_1}=H_{t_2,k_2}H_{k_2,t_1}$ and $H_{t_2, k_1}=H_{t_2,k_2}H_{k_2,k_1}$. The calculation of $H_{t_3,k_2}$ is similar to the previous criterion.
 \end{itemize}

\end{enumerate}

{\color{red}{The aforementioned computation process involves only interpolation calculations and matrix multiplication, which ensures high computational efficiency in practical implementation. Through the derivations described above, we can rapidly and cost-effectively calculate the homography matrix between any two frames. Specifically, the parameter $h$ directly controls the balance: 1) Higher Accuracy (Smaller $h$), a smaller $h$ means more frequent direct calculations, leading to higher accuracy but increased computational cost; 2) Higher Efficiency (Larger $h$), A larger $h$ significantly reduces computational load, improving speed (FPS) while accepting a minor approximation.}}

\subsection{View-Centric ID Learning}

Building upon the previously calculated homography matrix between adjacent frames, we can facilitate multi-view learning of object IDs.

In MOT, ID learning is critical for identifying object ID information to facilitate data association. Due to the dynamic nature of UAV-MOT scenes where both the background and objects may be moving, the objects may exhibit a rich variety of view changes. Therefore, we propose a novel View-Centric ID Learning  approach, which is inspired by the Object-Centric Learning framework \cite{slot}. Specifically, in combination with the homography of the view plane in different frames, we first propose homography Slot Attention, which uses "slots" \cite{slot,slotdiffusion} to explicitly learn the view information of an object, referred to as View-Centric Slots. We correlate the extracted view features of objects from adjacent frames to obtain updated view slots, and then through a decoding process, we reconstruct a new ID embedding and update the object's ID embedding feature.

Within the foundational concept of Slot Attention, The Slot Attention \cite{slot, slotdiffusion} maps from a set of input feature vectors to a set of output vectors, referred to as slots. Each vector in this output set can describe an object or an entity in the input.  Building on this concept, we propose Homographic Slot Attention (HSA), which leverages the view-based homography matrix and iterative attention to map its inputs to view-centric slots. At each iteration, slots compete to describe views of the input through a softmax-based attention mechanism and homography matrix. As shown in Figure \ref{fig:vcil}, given two frames $t, t+1$, we describe their one iteration of HSA on the ID feature extracted by the detection head as,

\begin{equation}
\begin{aligned}
    {\rm HSA}_{t} &= {\rm Softmax}(\frac{k(F_{t}^{ID}) (H_{t, t+1}  q(S_{t})^{\top})}{\sqrt{C}})  v(F_{t}^{ID}), \\
    {\rm HSA}_{t+1} &= {\rm Softmax}(\frac{k(F_{t+1}^{ID})  (H_{t+1, t}  q(S_{t+1})^{\top})}{\sqrt{C}} ) v(F_{t+1}^{ID}),
\end{aligned}
\end{equation}

\noindent where $q(\cdot), k(\cdot), v(\cdot)$ are the linear projections respectively. $F_{t}^{ID}, F_{t+1}^{ID}$ are the object ID features in frame $t, t+1$, respectively. $S_{t}, S_{t+1}$ are the View-Centric Slots in frame $t, t+1$, respectively. $\top$ is the matrix transposition operator. $C$ is the feature dimension. Then we correlate the cross-frame views by,
\begin{equation}
    S' = {\rm Softmax}(\frac{S_{t}  S_{t+1}^{\top}}{\sqrt{C}})   S_{t+1} 
\end{equation}
\noindent then a MLP-decoder $\mathcal{D}$ is used to reconstruct visual  feature ${F^{ID}}'$ from the correlated slots $S'$.  Finally,  ${F^{ID}}'$ is correlated with $F_{t_2}^{ID}$(current frame) for ID feature updating, in the cross-attention manner.

\subsection{Homography Matching Filter}

Leveraging the homography matrix, we design a Homographic Matching Filter (HMF) to map the object bounding boxes from different frames onto the same view plane, allowing for more accurate physical IoU associations that closely mirror the real world.

In UAV video sequences, the movement of objects is not limited to linear motion but becomes nonlinear and irregular due to the coupling of the UAV's and the object's own motions. The traditional Kalman filter \cite{kalman} struggles with this irregular motion, so we introduce the HMF to handle the complex movements associated with UAVs. The HMF module adaptively projects cross-frame boxes onto the same view plane, making IoU association more regular and normalized, which allows for accurately completing the object ID association, relative to the real world.

\begin{figure}
\centering
    \includegraphics[width=0.4\textwidth]{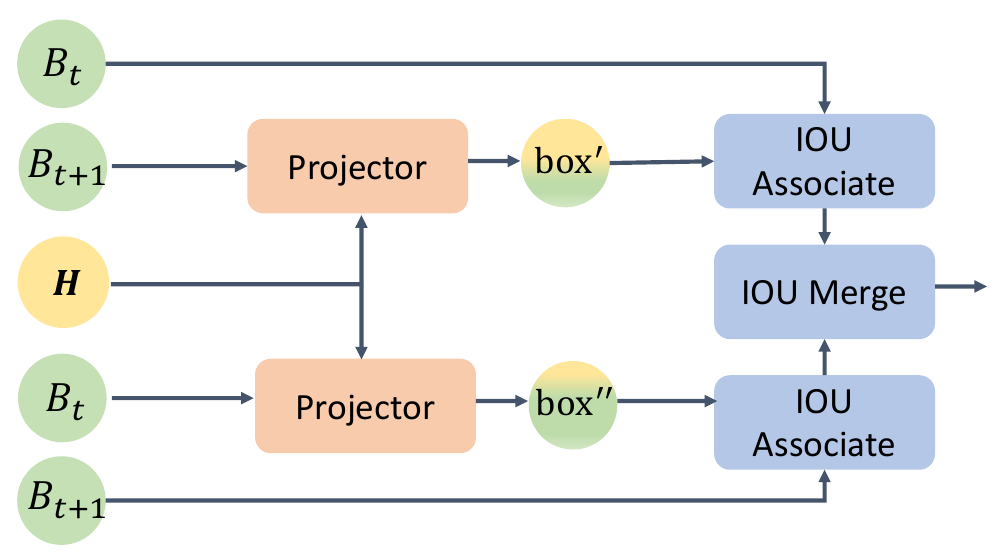}
    \caption{
       The illustration of Homography Matching Filter (HMF). We apply Homographic mapping between two adjacent frames and compute the Homographic IoU for each, then take the average of the IoUs to enhance robustness.
    } \label{fig:IoU_match}
  % \end{minipage}
\end{figure}

As shown in Figure \ref{fig:IoU_match}, to ensure robustness, we perform cross-projection between the object bounding boxes $\{B_{t}\}$ and $\{B_{t+1}\}$ of two adjacent frames $t, t+1$. For each pair of  $B_{t}$ and $B_{t+1}$, the calculation of HMF is,

\begin{equation}
\begin{aligned}
    {\rm HMF}(B_{t}, B_{t+1}) = (&{\rm IoU}(H_{t, t+1} B_{t}, B_{t+1})+ \\
     &{\rm IoU}(H_{t+1, t} B_{t+1}, B_{t})) / 2
\end{aligned}
\end{equation}

\noindent\textbf{Comparison with existing Motion Filters.} Previous motion filters in UAV MOT, such as Adaptive Motion Filter (AMF) in \cite{uavmot}, typically rely on the spatial relationships of objects to their surroundings to determine the UAV's movement state. However, this approach is heavily dependent on manual design and lacks robustness when the UAV's motion state is complex, as well as obviously affect the tracking efficiency. In contrast, our proposed HMF is more concise, lightweight, and robust.

\begin{algorithm}[!htb] % htbp 是推荐的位置参数
\caption{HomView-MOT Algorithm}
\label{algo}
\small
\begin{algorithmic}[1] % 数字表示行号
\Require An UAV video sequence $\{I_t \in \mathbb{R}^{W \times H \times 3}\}_{t=1}^T$
\Ensure The tracked objects $T_t = \{B_t, C_t, ID_t\}$
\While{$t < T$}
    \If{$t$ is divisible by $h$}
        \State Calculate $H_{t, t+h}$ of $I_t$, $I_{t+h}$ directly by Homography Estimation.
    \EndIf
    \State Derive $H_{t, t+1}$ from $H_{t, t+h}$ with Fast Homography Estimation algorithm.
    \State Input two adjacent frames $I_t$, $I_{t+1}$.
    \State Finish object detection, obtain the detected boxes $\{B_t, C_t\}$, $\{B_{t+1}, C_{t+1}\}$.
    \State Calculate IoU association with Homography Matching Filter.
    \State Obtain ID embedding features $F^{ID}_t$, $F^{ID}_{t+1}$, calculate the feature association.
    \State Update the ID feature $F^{ID}_{t+1}$ with View-Centric ID Learning.
    \State Ensemble IoU and feature association, perform Matching Algorithm.
\EndWhile
\end{algorithmic}
\end{algorithm}

\subsection{Loss Design and Overall Tracking Algorithm}

Along with the basic detection loss, we also add the reconstruction loss, for reconstructing the view-centric slots to visual feature. So the overall loss is formulated as, 
\begin{equation}
    \mathcal{L} = \mathcal{L}_{{\rm basic}} + \lambda \mathcal{L}_{rec} = \mathcal{L}_{{\rm basic}} + \lambda \sum_{t=1}^T\mathcal{L}_{CE}(F^{ID}_t, \mathcal{D}(S_t)),
\end{equation}
\noindent where $\lambda$ is loss weight. We use  cross-entropy loss both in  $\mathcal{L}_{rec}$.
The proposed tracking algorithm can be summarized as Algorithm \ref{algo}.

\section{Experiments}

\subsection{Datasets} \label{sec:dataset}

\subsubsection{VisDrone2019}: The VisDrone2019 dataset \cite{visdrone} is applied for tracking and detection from UAV perspectives. For the MOT task, the VisDrone2019 dataset comprises a training set (56 sequences), a validation set (7 sequences), and a test set (33 sequences, with test-challenge containing 16 sequences and test-dev comprising 17 sequences). In each frame, every object is labeled with a bounding box, category, and tracking ID. The VisDrone2019 dataset encompasses ten categories: pedestrian, person, car, van, bus, truck, motor, bicycle, awning-tricycle, and tricycle. For the multi-object tracking evaluation, only five object categories are considered, namely car, bus, truck, pedestrian, and van.

\subsubsection{UAVDT}: The UAVDT dataset \cite{uavdt} is specifically utilized for vehicle object detection and tracking, featuring three categories: car, truck, and bus. For the MOT task, it is divided into a training set (30 sequences) and a test set (20 sequences). It exclusively focuses on a single category, car. The video frames have a resolution of 1080 × 540 pixels and encompass various common scenes, such as squares, arterial streets, and toll stations.

\subsection{Implementation Details}

To ensure a fair comparison, we follow the experimental setup of existing works such as UAVMOT \cite{uavmot}. Regarding training details, the total number of epochs is set to 30, with the learning rate decaying by a factor of 10 at 10 epochs and 20 epochs, respectively. The initial learning rate is set to 7e-5. For data augmentation, we utilize random cropping and random scaling (between 0.6 to 1.3). The experiments are conducted on two GeForce RTX GPUs with a batch size of 4. In the multiple loss functions, L1 loss is used to supervise the object's width and height. Cross-entropy loss and triplet loss are employed for handling the object ID. {\color{red} For all main experiments on the VisDrone and UAVDT datasets, the value of $h$ was uniformly set to 10.}
Additionally, we utilize the standard cross-entropy loss to supervise the object heatmaps and view slot reconstruction. The loss weight $\lambda$ is set to 0.4.

\begin{table*}[]
\centering
\caption{Comparison with state-of-the-arts on VisDrone \textit{test-dev} dataset.}
\scalebox{1}{\begin{tabular}{l|c|c|ccccccccc}
\toprule
Method & Pub\&Year                                                          & FPS$\uparrow$  & MOTA$\uparrow$(\%) & MOTP$\uparrow$(\%) & IDF1$\uparrow$(\%) & MT$\uparrow$  & ML$\downarrow$   & FP$\downarrow$    & FN$\downarrow$     & IDs$\downarrow$  & FM$\downarrow$   \\ \midrule
MOTDT \cite{motdt}  & ICME2018                                                             & -    & -0.8 &  68.5 &  21.6 & 87  & 1196 & 44548 & 185453 & 1437 & 3609 \\
SORT \cite{sort} & ICIP 2016                                                          & 23.5 & 14.0 & 73.2 & 38.0 & 506 & 545  & 80845 & 112954 & 3629 & 4838 \\
IoUT \cite{IoUt} & AVSS2017                                                            & 27.3 & 28.1 & 74.4 &  38.9 & 467 & 670  & 36158 & 126549 & 2393 & 3829 \\
% GOG  \cite{gog}  & -                                                           & 2.0  & 28.7 & 76.1 &  36.4 & 346 & 836  & 17706 & 144657 & 1387 & 2237 \\
MOTR \cite{motr}  & ECCV2022                                                          & 7.5  & 22.8 & 72.8 &  41.4 & 272 & 825  & 28407 & 147937 & 959  & 3980 \\
TrackFormer \cite{trackformer} & CVPR2022                                                 & 7.4  & 25.0 & 73.9 & 30.5 & 385 & 770  & 25856 & 141526 & 4840 & 4855 \\
UAVMOT  \cite{uavmot} & CVPR2022                                                      & 12.0 & 36.1 & 74.2 & 51.0 & 520 & 574  & 27983 & 115925 & 2775 & 7396 \\
ByteTrack \cite{bytetrack} & ECCV2022 & 11.4 & 52.3 & -  & 68.3 & 769   & 325   & 24156   & 110342     & 2230    & 4001    \\
OC-SORT \cite{ocsort} & CVPR2023 & - &  39.6 & 73.3 & 50.4 & - & - & 14631 & 123513 & 986 & - \\
FOLT \cite{folt} & MM2023 & - &  42.1 & 77.6 & 56.9 & - & - & 24105 & 107630 & 800 & - \\
GLOA \cite{gloa} & J-STARS2023 & - &  39.1 & 76.1 & 581 & 824 & 18715 & 158043 & 4426 & - \\
U2MOT \cite{u2mot} & ICCV2023 &  19.4 & 52.3 & - & 69.0 & - & -  & - & -  & 1052 & -\\
STDFormer \cite{stdformer} & TCSVT2023 & - & 45.9 & 77.9 & 57.1 & 684 & 538 & 21288 & 101506 & 1440 & -\\
DroneMOT \cite{dronemot} & ICRA2024 & - &  43.7 & 71.4 & 58.6 & 689 & 397 & 41998 & 86177 & 1112 & - \\
AHOR-ReID \cite{ahor} & TCSVT2024 & 28.1&  42.5 & - & 56.4 & - & - & 21447 & 109762 & 810 & - \\
DFA-MOT \cite{dfa} & TCSVT2025 & 19.8 & 42.9 & - & 58.3 & 518 & 523 & 10760 & 119571 & 792 & -\\
\midrule
\textbf{HomView-MOT}   & -                                                                & 20.8 & \textbf{54.2} & 75.7 & \textbf{75.1} & \textbf{870} & \textbf{224}  & 31461 & 72600 & 1073  & \textbf{2832} \\ \bottomrule
\end{tabular}}
\label{table:visdrone}
\end{table*}

\subsection{Comparison with State-of-the-arts}

Following previous works\cite{uavmot, u2mot}, we benchmark our method against previous approaches on both the VisDrone2019 and UAVDT datasets for the MOT task. For VisDrone2019, we train our model using both the training and validation sets and assess our approach on the test-dev set utilizing the official VisDrone MOT toolkit. For UAVDT, we train the proposed network on the training set and evaluate on the test set.
As depicted in Table \ref{table:visdrone} and Table \ref{table:uavdt}, our method achieves significantly enhanced results compared to existing methods.

\subsection{Ablation Study}

\subsubsection{Ablation Study of Each Component}  We delve into each proposed component of HomView-MOT the show the results in Table \ref{table:abl}. Specifically, we evaluate HomView-MOT with different settings, including ``w/o FHE", ``w/o VCIL" and ``w/o HMF". Noted that, for the setting of  ``w/o FHE", we set the homography matrix of any two frames to be the homography matrix of adjacent key frames. As seen that all the components play an important role for the proposed tracking framework.

\begin{table*}[]
\centering
\caption{Comparison with state-of-the-arts on UAVDT \textit{test} dataset.}
\scalebox{1}{\begin{tabular}{l|c|ccccccccc}
\toprule
Method   & Pub\&Year                                                        & MOTA$\uparrow$(\%)  & MOTP$\uparrow$(\%)     & IDF1$\uparrow$(\%) & MT$\uparrow$  & ML$\downarrow$   & FP$\downarrow$    & FN$\downarrow$     & IDs$\downarrow$  & FM$\downarrow$   \\ \midrule
IOUT \cite{IoUt}    & AVSS2017                                                                  & 36.6  & 72.1  & 23.7 & 534 & 357  & 42245 & 163881 & 9938 & 10463 \\
SORT \cite{sort}    & ICIP2016                                                             & 39.0  & 74.3   & 43.7 & 484 & 400  & 33037 & 172628 & 2350 & 5787 \\
DSORT \cite{dsort}   & ICIP2017                                                                  & 40.7  & 73.2  & 58.2 & 595 & 338  & 44868 & 155290 & 2061 & 6432 \\
CenterTrack \cite{CenterTrack} & ECCV2020 & 40.9  & -  & 59.9 & 624 & 221  & 41681 & 159478 & 434 & 2093 \\
ByteTrack \cite{bytetrack}  & ECCV2020 & 41.6  & -  & 59.1 & - & -  & 28819 & 189197 & 296 & - \\
FOLT \cite{folt}  & MM2023 & 41.6  & -  & 48.5 & - & 68.3  & 36429 & 155696 & 338 & - \\
OCSORT \cite{ocsort}  & CVPR2023 & 47.5  & -  & 64.9 & - & 68.3  & 47681 & 148378 & 288 & - \\
UAVMOT \cite{uavmot}  & ICCV 2023 & 46.4 & 72.7 & 67.3 & 624  & 221 & 66352 & 115940 & 456 & 5590 \\
DistMOT \cite{DistMOT} & ICSIDP2024 & 40.4  & -  & 57.2 & - & -  & - & - & 593 & - \\
DFA-MOT \cite{dfa} & TCSVT2025 & 48.0 & - & 69.3 & 684 & 230 & 59218 & 119248 & 396 & 1269\\
\midrule

\textbf{HomView-MOT}    & -                                                                 & \textbf{48.1} & \textbf{74.3}  & \textbf{70.7} & \textbf{694} & 234  & 58834 & 117691 & 427 & 6003   \\ \bottomrule
\end{tabular}}
\label{table:uavdt}
\end{table*}

\begin{table}[]
\centering
\caption{Ablation study of each component on VisDrone.}
\scalebox{1}{\begin{tabular}{ccc|ccc}
\toprule
 FHE & VCIL & HMF & MOTA $\uparrow$ (\%) & IDF1 $\uparrow$ (\%) & IDs $\downarrow$  \\ \midrule
   $\checkmark$  &   $\checkmark$   &  $\checkmark$   &  54.2  & 75.1 & 1073   \\
      &    $\checkmark$  &  $\checkmark$   &   42.2   &  64.7   &  2350    \\
    $\checkmark$   &      &  $\checkmark$   &   48.8   &  68.5   &  1566    \\
    $\checkmark$   &   $\checkmark$   &    &    50.0   &   70.9  &   1322    \\
    \bottomrule
\end{tabular}}
\label{table:abl}
\end{table}

{\color{red}\subsubsection{Ablation Study on VCIL}  View-Centric ID Learning (VCIL) enables ID embedding to learn multi-view information, effectively enhancing the reliability of ID feature association and thereby bolstering the robustness of MOT algorithms in moving UAV scenarios. To verify the effectiveness of VCIL, we study its benefits on ID feature association using two metrics: IDs, IDF1. For fair comparison with previous methods, we follow the same baseline with \cite{uavmot}. Specifically, we evaluate the performance of the baseline, the baseline with the IDFU module introduced in \cite{uavmot}, and the baseline with VCIL. The experimental results, as shown in Table \ref{table:vcid}, indicate that the inclusion of VCIL leads to a significant improvement across all  metrics compared to both the baseline and IDFU.}

{\color{red}\subsubsection{Comparison between VCIL and ReID methods}  To quantitatively validate VCIL’s advantages, we compare it against two standard ReID-based MOT baselines on the VisDrone dataset: 1) DeepSORT-style ReID, uses a pre-trained ReID model for feature extraction; 2) FairMOT-style Joint Detection and ReID,  integrates ReID learning into the detection head. The results shown in Table. \ref{table:vcid_reid} demonstrate VCIL’s superiority in handling view changes specific to UAV scenarios. VCIL significantly reduces identity switches (IDs) and improves IDF1, highlighting its ability to maintain ID consistency under challenging viewpoint variations where standard ReID methods fail.}

\begin{figure}
    \centering
    \includegraphics[width=0.7\linewidth]{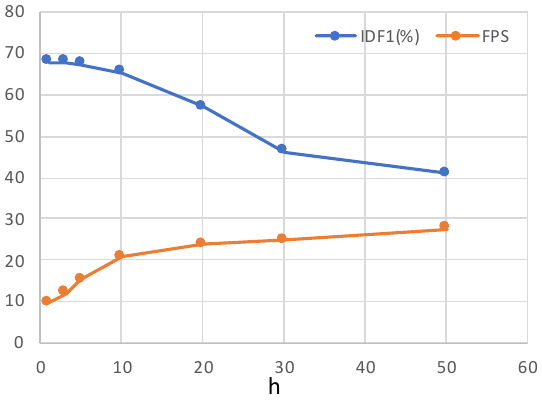}
    \caption{
       Ablation Study on $h$ in FHE on VisDrone. 
    } \label{exp:robust}
\end{figure}

{\color{red}\subsubsection{Ablation Study on $h$ in FHE} In order to ensure the overall efficiency of tracking, we compute the inter-frame homography Matrix rapidly by using interval sampling coupled with FHE, based on the assumption that the scene can be reasonably considered to be moving linearly within a small sampling interval. This design ensures the efficiency and precision of our HomView-MOT. Generally, a larger sampling interval tends to increase tracking efficiency, but an overly large interval may weaken the assumption of "approximate linear motion within the interval," thus compromising tracking accuracy. Figure \ref{exp:robust} displays the performance of tracking under different settings of \( h \). It is observed that a smaller \( h \) yields the highest accuracy but also reduces speed. As \( h \) increases, we maintain a high level of accuracy while also achieving a higher speed, but a further increase in \( h \) will lead to a noticeable decrease in accuracy. These experimental findings strongly support the aforementioned arguments. Specifically, When $h=1$,  FHE degenerates into the ordinary HE algorithm, yielding high accuracy but at a slow speed. As $h$ increases, the tracker maintains relatively high accuracy while the speed significantly improves, confirming the value of the FHE approach. However, if $h$ is too large, the calculation of the homography matrix becomes imprecise, which affects performance. Therefore, in all the experiments, we set $h=10$ to balance the accuracy and efficiency. }

\begin{table}
\centering
    \caption{Ablation study on VCIL on VisDrone.}
    \scalebox{1}{\begin{tabular}{c|cc}
    \toprule
                  & ~ IDs$\downarrow$~ & IDF1$\uparrow$(\%)   \\ \midrule
    Same Baseline as  \cite{uavmot}    &  2079   &  40.6             \\
    +IDFU\cite{uavmot} ~  & 937 & 43.8             \\ \midrule
    \textbf{+VCIL} &  \textbf{790}  &   \textbf{51.9}     \\ \bottomrule
    \end{tabular}}
    \label{table:vcid}
\end{table}

{\color{red}\subsubsection{Comparison between HMF and the existing motion filters } 

Motion-aware filters such as  Kalman excel at modeling object dynamics because the camera’s perspective is stable. However, in moving-UAV scenarios, the camera’s ego-motion (e.g., abrupt turns, altitude changes) often dominates the observed motion of objects. This makes it difficult to disentangle an object’s true motion from the camera’s movement using traditional motion models alone. The proposed HMF is to first eliminate the dominant camera motion via homography projection, thereby recovering the object’s residual motion in a common reference plane. As shown in Table \ref{table:hmf}, the Kalman filter struggles with erratic UAV ego-motion, leading to frequent trajectory mismatches (high IDs). In contrast, HMF’s homography-based stabilization proves more robust to camera dynamics, achieving higher accuracy and fewer identity switches. This validates that in moving-UAV settings, compensating for camera motion geometrically (via HMF) is more effective than predicting object motion under unstable camera dynamics.}

\begin{table}
\centering
    \caption{Comparison between VCIL and ReID methods on  VisDrone.}
    \scalebox{1}{\begin{tabular}{c|cc}
    \toprule
              &  ~ IDs$\downarrow$~ & IDF1$\uparrow$(\%)   \\ \midrule
    w/ DeepSORT ReID \cite{deepsort}    &  2911   &  38.3             \\
    w/ FairMOT-style ReID \cite{fairmot} ~  & 1033 & 41.7             \\ \midrule
    \textbf{w/ VCIL} &  \textbf{790}  &   \textbf{51.9}     \\ \bottomrule
    \end{tabular}}
    \label{table:vcid_reid}
\end{table}

\begin{table}[]
    \centering
    \caption{The Closer Observation on HMF on VisDrone.}
    \scalebox{1}{\begin{tabular}{c|cc}
    \toprule
                 & ~ IDs$\downarrow$ & ~ IDF1$\uparrow$(\%)    \\ \midrule
    Same Baseline as  \cite{uavmot}     &   2079  & 40.6                   \\
    +AMF\cite{uavmot} ~ &  958   &  44.1                       \\ \midrule
    \textbf{+HMF} &  \textbf{816}    &   \textbf{53.7}                     \\ \bottomrule
    \end{tabular}}
    \label{table:hmf}
\end{table}

\subsubsection{Case Study} 
\begin{figure}
    \centering
    \includegraphics[width=1\linewidth]{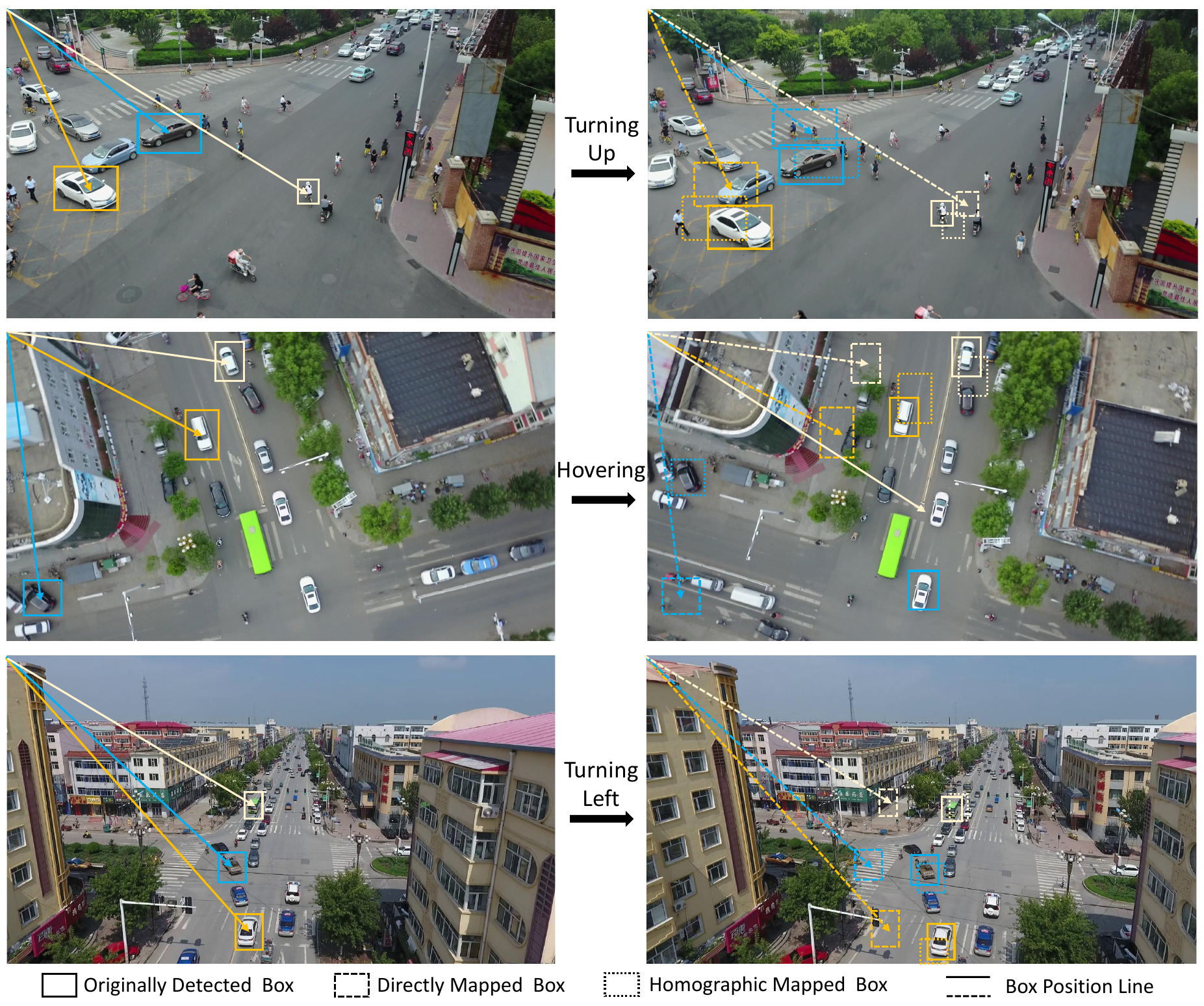}
    \caption{The case studies of HomView-MOT. We show three UAV flight states: turning up, Hovering in the sky, and turning left. As seen that the proposed Homographic Matching Filter demonstrate better and more reasonable IoU association (relative to the real world) then traditional ordinary IoU association, by comparing the Directly Detected Boxes and Homographic Mapped Boxes.}
    \label{fig:case}
\end{figure}

To better illustrate the effectiveness of the HMF, we visually present in Figure \ref{fig:case} the performance of HMF when the UAV-captured scene moves, causing the target trajectories to become irregular. Figure \ref{fig:case} depict three different UAV status changes: hovering, turning up, and turning left, showing comparison between two consecutive frames. 
It is evident that the objects have not moved significantly in the real world (as can be seen with reference to ground features such as lane lines), and when directly mapped, the IoU between the detection boxes of consecutive frames is very low, leading to potential match losses. However, with the mapping based on homography, the detection boxes from frame to frame are closer together, yielding a higher IoU and thus achieving a more reliable IoU association.

{\color{red}\subsubsection{Discussion of the generalization capability} The HomView-MOT framework is fundamentally designed to handle challenges that are universal across moving-camera scenarios, which naturally provides a degree of robustness against domain shifts. The HMF does not rely on learning dataset-specific motion patterns. Instead, it uses scene homography, a geometric property, to project object bounding boxes into a common view for association. 
Therefore, even if the UAV's flight pattern (e.g., more aggressive turns) or the environment (e.g., from an urban grid to a highway) changes significantly, the underlying geometric principle remains valid. This makes HMF less sensitive to domain shifts in motion dynamics compared to Kalman-filter-based approaches that learn from training data. }

\section{Conclusion}

We propose the HomView-MOT framework that successfully addresses the inherent challenges of multi-object tracking in moving UAV scenarios. By employing the principles of homography and view-centric learning, our approach effectively navigates the complexities introduced by UAV flight dynamics. The FHE algorithm computes homography matrices efficiently, while our HMF and VCIL techniques facilitate accurate and robust tracking across changing scenes. 
The effectiveness of our framework is underscored by its state-of-the-art performance on the VisDrone and UAVDT datasets.

{\color{red}\textbf{Limitation.} The core design of HomView-MOT is not exclusively limited to UAV scenarios. The VCIL and HMF modules are designed to address the fundamental issue of scene motion, which is a common challenge in all moving-camera tracking. However, performance gains may vary in non-UAV contexts. In scenarios with milder and more predictable homography changes, the performance improvement offered by the HMF might be less dramatic than that in highly dynamic UAV flights, as the standard IoU association may already be reasonably effective. Nevertheless, the VCIL module's ability to robustly handle view changes could still contribute to the tracking robustness.}

\bibliographystyle{unsrt}
\bibliography{custom}

\end{document}